\documentclass[letterpaper]{article} 
\usepackage{aaai25}  
\usepackage{times}  
\usepackage{helvet}  
\usepackage{courier}  
\usepackage[hyphens]{url}  
\usepackage{graphicx} 
\usepackage{natbib}  
\usepackage{caption} 
\usepackage{algorithm}
\usepackage{algorithmic}

\usepackage{newfloat}
\usepackage{listings}
\DeclareCaptionStyle{ruled}{labelfont=normalfont,labelsep=colon,strut=off} 
\floatstyle{ruled}
\newfloat{listing}{tb}{lst}{}
\floatname{listing}{Listing}
\usepackage{multirow}
\usepackage{amsmath}
\usepackage{amssymb}

\title{ST-ReP: Learning Predictive Representations Efficiently for Spatial-Temporal Forecasting}
\author{
    Qi Zheng\textsuperscript{\rm 1},
    Zihao Yao\textsuperscript{\rm 1},
    Yaying Zhang\textsuperscript{\rm 1}\thanks{Corresponding author.}
}
\affiliations{
    \textsuperscript{\rm 1}The Key Laboratory of Embedded System and Service Computing, Ministry of Education, \\Tongji University, Shanghai 200092, China\\
    \{zhengqi97, yaozihao, yaying.zhang\}@tongji.edu.cn
}

\begin{document}

\maketitle

\begin{abstract}
Spatial-temporal forecasting is crucial and widely applicable in various domains such as traffic, energy, and climate. Benefiting from the abundance of unlabeled spatial-temporal data, self-supervised methods are increasingly adapted to learn spatial-temporal representations. However, it encounters three key challenges: 1) the difficulty in selecting reliable negative pairs due to the homogeneity of variables, hindering contrastive learning methods; 2) overlooking spatial correlations across variables over time; 3) limitations of efficiency and scalability in existing self-supervised learning methods. To tackle these, we propose a lightweight representation-learning model ST-ReP, integrating current value reconstruction and future value prediction into the pre-training framework for spatial-temporal forecasting. And we design a new spatial-temporal encoder to model fine-grained relationships. Moreover, multi-time scale analysis is incorporated into the self-supervised loss to enhance predictive capability. Experimental results across diverse domains demonstrate that the proposed model surpasses pre-training-based baselines, showcasing its ability to learn compact and semantically enriched representations while exhibiting superior scalability.
\end{abstract}

%
\begin{links}
    \link{Code}{https://github.com/zhuoshu/ST-ReP} 
\end{links}

\section{Introduction}
\begin{figure}[t]
	\centering
	\includegraphics[width=\columnwidth]{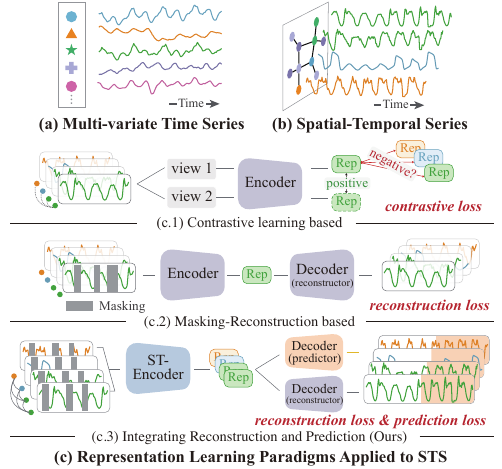}
	
        \caption{Top: An illustration of typical multivariate series and spatial-temporal series. Different shapes denote types of variables. STS usually has homogeneous ones. Bottom: Comparison of three examples of representation learning paradigms. Our proposed model belongs to the third one.}

        \label{fig_intro}
\end{figure}
Spatial-temporal (ST) forecasting, a crucial task within multivariate time series (MTS) analysis, plays an essential role in understanding the operational dynamics of urban and geospatial systems. With the rapid expansion of data monitoring technologies, there is an unprecedented increase in unlabeled spatial-temporal series (STS) data, prompting growing interest in self-supervised learning approaches. These approaches exploit large-scale data to generate pseudo-labels for training models to effectively represent raw data. Although self-supervised learning has shown promise in MTS modeling through contrastive and reconstruction-based methods, its application to STS data is challenged by variable homogeneity and the inherent complexity of spatial-temporal attributes:

\textbf{1) Crafting suitable positive and negative pairs for contrastive methods in the context of STS is challenging due to the variable homogeneity.} As depicted in Figure \ref{fig_intro}(a)(b), in MTS, variables often represent different attributes (such as oil temperature, external load, and location features of a power transformer). When comparing sequences from distinct attributes, it is possible to learn features that exhibit clear differences. In contrast, STS variables, typically from similar sources like traffic network's multiple road sensors, possess homogeneity when measuring the same data type, making variable distinction challenging. 
When contrastive methods treat representations of different variables within the same batch as negative samples, as shown in Figure \ref{fig_intro}(c.1), there is a significant risk of generating false negatives. This occurs because these variables, despite being considered negatives, may actually be closely related or exhibit similar patterns in the spatial-temporal context, thereby hindering the accurate extraction of valuable information.

\textbf{2) Existing self-supervised methods often overlook spatial correlations between variables.} As shown in Figure \ref{fig_intro}(c.2), many reconstruction-based approaches aim to enhance model understanding and representation of MTS by reconstructing input data. These methods generally focus on individual variables and extract features predominantly from the temporal dimension, with limited consideration of inter-variable relationships. However, an important characteristic of STS data is the spatial correlation between variables, which has been frequently studied in recent years. Ignoring this correlation can lead to insufficient spatial-temporal extraction.

\textbf{3) The efficiency and scalability of self-supervised methods demand augmentation.} As illustrated in Figure \ref{fig_intro}(c.3), unlike MTS models that treat different variable sequences as distinct samples, STS models primarily consider all variable sequences within a given time period as a single sample to account for their potential spatial correlations. However, this approach can increase the computational burden during training, particularly as the number of nodes grows. In most high-complexity algorithms, the computational cost escalates significantly. Thus, efficient and effective self-supervised models for practical spatial-temporal forecasting needs remain an area for further research.

To address these challenges, we propose a reconstruction-based self-supervised model \textit{ST-ReP} to learn representations for \underline{\textbf{S}}patial-\underline{\textbf{T}}emporal forecasting. It integrates \underline{\textbf{Re}}construction with \underline{\textbf{P}}rediction, explicitly considering spatial correlations between variables during the encoding process, as illustrated in Figure \ref{fig_intro}(c.3).
Specifically, we extend the reconstruction objective to encompass not only the current series but also unseen future values, thereby encouraging the model to learn representations that have predictive capabilities. Additionally, we propose a new ST encoder, to model the spatial-temporal relationships between variables with compression and decompression. It first reduces the temporal dimension to a compact latent space, then employs a linear encoder to compute spatial correlations between variables, and finally restores the original temporal length. This approach aims to reduce redundancy in extracting spatial-temporal features while also lowering computational complexity.

Moreover, we introduce a loss function to capture the multi-scale temporal information in the self-supervised training process, which links the reconstructed values and the predicted ones. This method constrains the model to understand temporal correlations from multiple perspectives without increasing the computational cost during encoding. With these designs on pre-training framework and the encoder mechanism, our proposed model ST-ReP is capable of learning semantically dense and predictive ST representations efficiently. Our contribution are summarized as follows:
\begin{itemize}
    \item To capture unique spatial interactions inherent in spatial-temporal data, we propose a reconstruction-based representation learning method called ST-ReP. By combining the reconstruction of current series with the prediction of future values, ST-ReP is encouraged to learn spatial-temporal correlations and generate predictive representations.
    \item  We propose a spatial-temporal encoding scheme called \textit{Compression-Extraction-Decompression} (\textit{C-E-D}). It involves \textit{Compression-Decompression} in the temporal dimension to transform features into a compact latent space, with spatial feature \textit{Extraction} in between. Additionally, we utilize a linear encoder in the spatial dimension to capture correlations between variables, thereby reducing potential redundancy.
    \item Experiments are conducted on six spatial-temporal datasets from various domains. The results demonstrate that our model achieves superior downstream prediction accuracy compared to advanced self-supervised learning baselines. Furthermore, our model showcases lower footprints and learns representations with lower dimensions, highlighting its good scalability.
\end{itemize}

\section{Related Work}
In this section, we will review self-supervised representation learning methods and their adaptation in STS data. Due to strong generalization and adaptability, the self-supervised learning method has garnered significant attention in MTS modeling \cite{survey_ssl4ts}. Effective self-supervised learning relies on a suitable encoder, clear pseudo-labeling principles, and robust loss constraints, and typically falls into two main categories: contrastive methods and generative methods.

\paragraph{Constrastive-based methods}
Contrastive methods perform pseudo-classification tasks in the representation space. They primarily rely on data augmentation and the design of positive and negative pairs to make the learned representations as distinguishable as possible. Numerous studies for MTS adopt this paradigm \cite{TS2Vec,CoST,T-Rep,TimesURL}. They often select representations of other variables within the same batch as negative examples. However, due to the homogeneity of variables in STS data, this selection has inherent limitations and may provide misleading information. Then ST-SSL \cite{ST-SSL} introduces self-supervised learning into spatial-temporal forecasting tasks. It augments ST graphs for capturing spatial-temporal heterogeneity under the contrastive setting. In contrast, our proposed ST-ReP does not adopt a contrastive strategy. It avoids multi-view augmentations that increase the encoding burden and sidesteps the potential inductive biases arising from unclear selection of positive and negative pairs.

\paragraph{Reconstruction-based methods}
Generative methods primarily include reconstruction-based approaches. These methods perform regression tasks in the original input space, typically using an auto-encoder framework to restore original data from masked visible inputs. This approach forces the model to understand underlying patterns in the data and enhances its robustness. Many methods adopt similar strategies for representation learning \cite{TST,PatchTST,SimMTM}. These methods primarily model individual time series without considering spatial correlations between variables. STEP \cite{STEP} and GPT-ST \cite{GPTST} are two recent studies that introduce self-supervised learning into STS forecasting. However, their focus is primarily on using self-supervised learning as an auxiliary tool to enhance the performance of complex downstream ST models. They do not specifically evaluate the spatial-temporal representations learned by the pre-training models.

Moreover, most methods use highly complex encoding structures or high-dimensional representations under multi-view augmentation, leading to high computational overhead, which becomes a bottleneck when dealing with large-scale datasets with more than thousands of variables. By comparison, our ST-ReP introduces completely unseen future values as partly reconstruction targets to enhance the predictive performance of the representations. It captures correlations between variables while maintaining high encoding efficiency and scalability.

\begin{figure}[t]
	\centering
	\includegraphics[width=\linewidth]{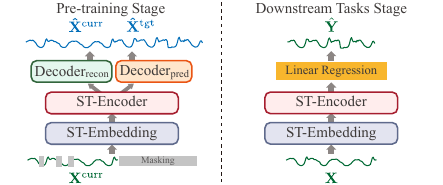}
	\caption{The overall workflow of ST-ReP.}
        \label{fig_workflow}
\end{figure}

\begin{figure*}[t]
	\centering
	\includegraphics[width=\linewidth]{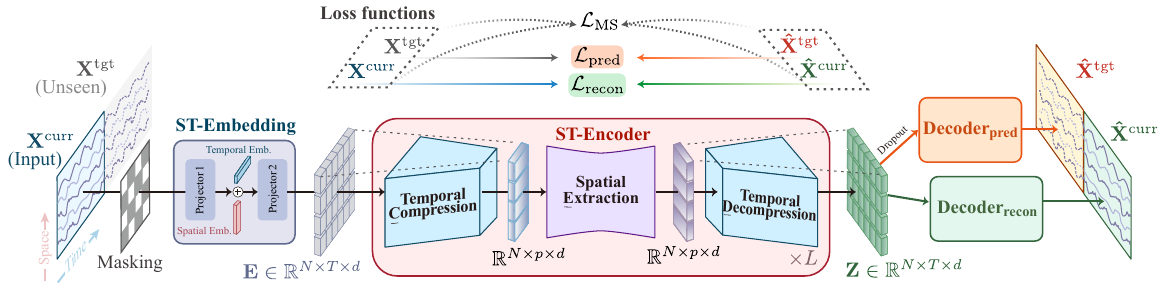}
	\caption{The overall pre-training framework of our proposed ST-ReP. The ST-Encoder utilizes a Compression-Extraction-Decompression structure to enhance modeling efficiency. ST-ReP integrates reconstruction and prediction, employing three types of loss functions to supervise representation learning: reconstruction loss $\mathcal L_{\text{recon}}$, prediction loss $\mathcal L_{\text{pred}}$, and multi-scale loss $\mathcal L_{\text{MS}}$.}
        \label{fig_model}
\end{figure*}
\section{Preliminary}
A set of STS data with $N$ variables can be formulated as a three-dimensional tensor $\mathbf X \in \mathbb {R}^{N\times T \times C}$ in the previous $T$ time intervals, and $C$ is the number of features of each variable. In this work, rather than focusing on end-to-end prediction, we aim to learn ST representations that can extract information beneficial for downstream prediction tasks. Formally, we need to learn a function $f$ that maps the raw data to a $d$-dimensional representation space: 
$\{r_1,...,r_i,...,r_N\}=f(\{X_1,...X_i,...,X_N\})$.
Here, $r_i \in\mathbb R^{T \times d}$ represents the representation of variable $i$, with $r_{i,t_j}\in \mathbb{R}^d $ being the representation vector at time step $t_j$. Once representations of past data are obtained, they can be used for various downstream tasks (such as forecasting and imputation). In this paper, we focus on ST forecasting, which involves using data from the past $T$ time steps to predict data for the future $T'$ time steps.

The primary goal of self-supervised learning is to benefit from pre-training on abundant data, enabling the model to understand the data and convert raw inputs into valuable representations. To assess the quality of these learned representations, we employ simple and lightweight downstream models (e.g. linear regression) instead of relying on complex models (e.g. deep learning methods), thereby demonstrating the predictive performance in the most basic scenarios. The overall workflow of our method is shown in Figure \ref{fig_workflow}. This approach also showcases the adaptability of self-supervised learning in resource-constrained environments. 

\section{Methodology}
The core idea of our proposed ST-ReP is to use current ST series data $\mathbf X^{\text{curr}} \in \mathbb R^{N \times T \times C}$ by masking and reconstructing it, while simultaneously predicting future series values $\mathbf X^{\text{tgt}} \in \mathbb R^{N \times F \times C}$. $F$ is the predicted length of the targeted value in the pretext task. Figure \ref{fig_model} provides an overview of our model's architecture, which follows a self-supervised pretraining framework. Similar to many reconstruction-based methods \cite{STEP}, our model initially applies random masking to the current values $\mathbf X^{\text{curr}}$ along the temporal dimension (masking is only used during training), and then utilizes an embedding module and encoder to generate data representations. Differently, ST-ReP
prioritizes a lightweight encoder structure to minimize redundancy during feature extraction. Following encoding, ST-ReP employs a reconstructor ($Decoder_{recon}$) to restore the representations to their original values, and an independent predictor ($Decoder_{pred}$) to project them onto future values. That means $\mathbf X^{\text{tgt}}$ is only used to participate in the loss calculation and is totally unseen for encoder, which can be viewed as a complete mask for $\mathbf X^{\text{tgt}}$. Lastly, the model integrates the reconstructed current values and the predicted future values, performing multi-scale temporal comparison over long sequences, thereby compelling the model to learn temporal relationships among series elements. In the following subsections, we elaborate three key components in ST-ReP: the ST-Embedding module, the linear ST encoder with a \textit{C-E-D} structure, and the reconstruction and prediction integrated loss setting.

\subsection{ST-Embedding Module}
To capture both the temporal progression and inherent spatial characteristics of the data, we integrate these aspects using embedding techniques. 
Firstly, a projector maps the feature dimensions of the raw data into a hidden space:
\begin{equation}
    \mathbf {\widetilde{E}}^{\text{hidden}}=Projector_1(\mathbf {\widetilde {X}}^{\text{curr}}),
\end{equation}
where $\widetilde{\mathbf X}^{\text{curr}}$ is the masked input data. $Projector_1$ consists of two linear layers with a ReLU activation function in between. Following a recent study's proposed cross-time normalization strategy \cite{TAGnn}, we concatenate data at the most recent time step with the raw data as input to the projector. Next, we use learnable, randomly initialized mask tokens to fill in the hidden data to the original length and obtain $\mathbf {\widetilde{E}}^{\text{hidden}}\in \mathbb R^{N \times T \times d}$. We add three commonly-used embedding information: $E^{\text{tod}}$, $E^{\text{dow}}$, and $E^{\text{spt}}$. $E^{\text{tod}} \in \mathbb R^{T \times d}$, represents the time of day corresponding to each time interval (with an upper limit of $\mathcal T$), $E^{\text{dow}} \in \mathbb R^{T \times d}$ represents the day of the week (with an upper limit of $7$ days), and both are queried from learnable parameters of length $\mathcal T$ and $7$, respectively. Additionally, $E^{\text{spt}}$ represents the variable's index information. These embeddings are then added through broadcasting. Note that even in the masked parts, the temporal and spatial embeddings remain, reflecting the real-world scenario of data missing.
After incorporating static temporal and spatial identification information, we use another projector to enhance the semantic density of the series embedding:
\begin{equation}
    \mathbf E = Projector_2(\mathbf {\widetilde{E}}^{\text{hidden}}+E^{\text{tod}}+E^{\text{dow}}+E^{\text{spt}}).
    \label{equ_e}
\end{equation}
Here, a one-layer equal-width convolution along with the time dimension is employed as $Projector_2$. This results in an embedding $\mathbf E \in \mathbb R^{N \times T \times d}$ that combines spatial-temporal context.

\subsection{Compression-Extraction-Decompression structure for ST encoding}
The encoding is one of pivotal parts in ST representation learning, and various ST mining models can serve as the backbone for it. In this work, we focus on capturing cross-time spatial relationships among variables while emphasizing the use of linear complexity models for ST modeling. This approach aims to enhance encoding efficiency and scalability, enabling the model to handle STS with a large number of variables. Achieving this balance between reduced complexity and effective modeling is a challenging task, which has been increasingly studied in related research \cite{Informer,Pyraformer,Crossformer,Performer,Difformer}.

\begin{figure}[t]
    \centering
    \includegraphics[width=\columnwidth]{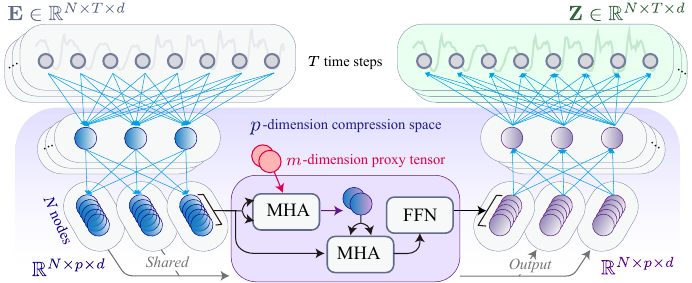}
    \caption{An example of the ST-Encoder. MLPs are used as the compressor and decompressor, with a linear spatail extractor based on proxy tensor.}
    \label{fig_STEncoder}
\end{figure}
To meet the objective, we propose a sandwich structure called \textit{Compression-Extraction-Decompression (C-E-D)} , as shown in Figure \ref{fig_model}. This structure compresses hidden features along the temporal dimension into a low-dimensional latent space, performs low-complexity spatial relationship extraction, and then restores the ST features to their original temporal length. Figure \ref{fig_STEncoder} presents an example that we used in this paper. We elaborate the details of each part as follows.

\paragraph{Temporal Compression and Decompression}
In this paper, we employ two fully connected layers with a non-linear activation function GELU \cite{GELU} in between, serving as the compressor and decompressor along the temporal dimension.
\begin{equation}
    \mathbf E^{comp}=Comp_{\text{tpr}}(\mathbf E) \in \mathbb R^{N\times p \times d},
    \label{equ_comp}
\end{equation}
\begin{equation}
    \mathbf H = Extraction_{\text{spt}}(\mathbf E^{comp}) \in \mathbb R^{N\times p \times d},
         \label{equ_extract}
\end{equation}
\begin{equation}
\mathbf E^{decomp}=Decomp_{\text{tpr}}(\mathbf H) \in \mathbb R^{N \times T \times d}.
\label{equ_decomp}
\end{equation}
The compressor maps the input embedding to a $p$-dimensional latent space, $p$ is a small constant value. The goal here is not to extract temporal correlations within the sequence but to summarize the current time series into a few key hidden states containing high-level semantics. This approach avoids potential redundancy and helps to reduce model overhead. Moreover, the decompression operation following spatial feature extraction (will be described in the next paragraph) is essential. It restores the low-dimensional hidden states to their original temporal length, providing an additional perspective to constrain the temporal semantics of the representation. This can be viewed as a form of temporal autoencoding. Additionally, this setup allows for the stacking of multiple layers, enhancing the model's expressive capacity.

\paragraph{Spatial Extraction}

We use the spatial node representations at a single time step as input tokens to compute attention among them, with the same encoding model shared across multiple time steps. Additionally, we employ a linear transformer structure to extract spatial correlations between variables, an area that has garnered increasing research interest \cite{Informer,Pyraformer,Crossformer,Performer,Difformer}. Specifically, we reduce the number of input tokens for attention calculation, similar to the router mechanism in \cite{Crossformer} and the bottleneck setup in \cite{SSTBAN}. Our approach involves using two multi-head attention layers along with a learnable tensor for proxy to capture spatial relationships effectively:

\begin{equation}
\begin{split}
    H^p &= \text{MHA}_1(P_w, E^{comp}, E^{comp}) \in \mathbb R^{m\times d}, \\
    H' &= \text{MHA}_2( E^{comp},H^p, H^p) + E^{comp} \in\mathbb R^{N \times d}, \\
     H &= \text{FFN}(H') +  H' \in\mathbb R^{N \times d}.
\end{split}
\end{equation}
For brevity, the subscript denoting the time dimension is omitted here, with unbolded symbols representing the hidden features at a single time step. Here, $P_w \in\mathbb R^{m \times d}$ is learnable parameters of constant size $m$. MHA (Multi-Head Attention) and FFN (Feed-Forward Network) are two fundamental components of the Transformer structure. The former takes three inputs as query, key, and value. $P_w$ acts as a proxy tensor in the attention computation, reflecting high-level semantics or summary information of the token sequence. This allows for an alternative interaction with the input tokens. Specifically, in $\text{MHA}_1$, $P_w$ serves as the query, while $E^{comp}$ serves as the key and value. The result $H^p$ is then used in $\text{MHA}_2$ as the key and value to restore the data features to the original token count. This setup significantly reduces computational overhead while maintaining expressive power, showing a linear computational complexity.

When capturing spatial correlations, each token already encapsulates the desired high-level temporal semantics. Consequently, the attention computation at this stage can effectively represent the extent of spatial interactions across time. After concatenating the spatial modeling results across $p$ virtual time steps, we obtain the final results of spatial extraction $\mathbf H \in \mathbb R^{N\times p\times d}$ for the succeeding decompression.
Equation \ref{equ_comp} to Equation \ref{equ_decomp} constitute one layer of the ST Encoder. Without loss of generality, the ST Encoder can be stacked in $L$ layers. We add the output of the final layer to the initial input $\mathbf E$ to obtain the final ST representation $\mathbf Z \in \mathbb R^{N \times T \times d}$.

\subsection{Reconstruction and Prediction integrated Loss Constraint}
After obtaining the representations, it is crucial to ensure they contain valuable semantic information. A key idea of ST-ReP is to combine reconstruction and prediction from generative methods. We use the learned representations $\mathbf Z$ from the masked input $\widetilde{\mathbf X}^{\text{curr}}$ to both reconstruct the current series $\mathbf X^{\text{curr}}$ and predict future values $\mathbf X^{\text{tgt}}$:
\begin{equation}
    \hat {\mathbf X}^{\text{curr}} = Decoder_{recon}(\mathbf Z) \in \mathbb R^{N \times T \times C},
\end{equation}
\begin{equation}
    \hat {\mathbf X}^{\text{tgt}} = Decoder_{pred}(Drop(\mathbf Z)) \in \mathbb R^{N \times F \times C}.
\end{equation}
Specifically, we use two linear layers with a GELU activation function in between as the $Decoder_{recon}$, with a residual connection before and after the first linear layer. For the $Decoder_{pred}$, we use two linear layers. $Drop$ represents the dropout operation, which is used to mitigate overfitting. These layers map the representations to the target dimensions in both the time and feature dimensions. Both the reconstruction and prediction tasks involve calculating the regression loss against the ground truth values:
\begin{equation}
    \mathcal L_{recon} = g(\hat {\mathbf X}^{\text{curr}},\mathbf X^{\text{curr}}), 
    \mathcal L_{pred} = g(\hat {\mathbf X}^{\text{tgt}},\mathbf X^{\text{tgt}}).
\end{equation}
where $g$ denotes the distance metric function, we use Huber loss \cite{HuberLoss} in this paper.

Furthermore, to encourage the model to acquire representations that retain as much temporal information as possible, we concatenate the predicted future values $\hat {\mathbf X}^{\text{tgt}}$ with the reconstructed current values $\hat {\mathbf X}^{\text{curr}}$, denoted as $\hat {\mathbf X}^{\text{full}}$. Subsequently, we compare this concatenated sequence $\hat {\mathbf X}^{\text{full}}$, with a length of $T+F$, to its corresponding ground truth values $\mathbf X^{\text{full}}$ across multiple time scales. Specifically, we utilize a set of kernels $\Omega$ with different size to conduct average pooling on both sequences separately, calculate the information loss at each scale, and then aggregate the results:
\begin{equation}
    \mathcal L_{MS} =\sum_{k\in \Omega}g(AvgPool_k(\hat {\mathbf X}^{\text{full}}),AvgPool_k(\mathbf X^{\text{full}}))
\end{equation}
This allows the model to capture more temporal granularity information and embed it within the learned ST representation. In contrast to many methods\cite{TimesNet,NHITS,MAGNN,Scaleformer,TimeMixer} that cascade to extract multi-scale information during encoding, we introduce multi-scale analysis solely into the loss computation, thus aiding in reducing model overhead.

Finally, the total loss is a linear combination of these three components:
\begin{equation}
    \mathcal L_{total} = \alpha \mathcal L_{recon} + \beta \mathcal L_{pred} + \gamma \mathcal L_{MS},
\end{equation}
where $\alpha$, $\beta$, and $\gamma=1-\alpha-\beta$ are weights of three parts.

\begin{table}[t]
\centering
\begin{tabular}{c|c|c|c|c}
\hline
\textbf{Dataset} & \textbf{\#Nodes} & \textbf{\#Samples} & \textbf{\begin{tabular}[c]{@{}c@{}}Time\\ Interval\end{tabular}} & \textbf{CV\textsuperscript{*}} \\
\hline
PEMS04          & 307  & 16992 & 5min   & 58.82 \\
PEMS08          & 170  & 17856 & 5min   & 46.75 \\
CA              & 8600 & 35040 & 15min  & 60.10 \\
SDWPF           & 134  & 35280 & 10min  & 121.97 \\
Humidity        & 2048 & 8784  & 1hour  & 17.19 \\
Temperature   & 2048 & 8784  & 1hour  & 2.19  \\
\hline
\multicolumn{5}{l}{* CV denotes the coefficient of variation.}
\end{tabular}

\caption{Summary of datasets.}
\label{table_dataset_details}
\end{table}

\section{Experiments}
We conduct experiments on datasets from diverse domains to evaluate the performance of our proposed ST-ReP.

\subsection{Experiment Setup}

\begin{table*}[ht]
\centering

\begin{tabular}{c|c|cc|cc|cc|cc|cc|cc}
\hline
\multicolumn{2}{c|}{\multirow{2}{*}{methods}}& \multicolumn{2}{c|}{\textbf{PEMS04}} & \multicolumn{2}{c|}{\textbf{PEMS08}} & \multicolumn{2}{c|}{\textbf{SDWPF}} & \multicolumn{2}{c|}{\textbf{Temperature}} & \multicolumn{2}{c|}{\textbf{Humidity}} & \multicolumn{2}{c}{\textbf{CA}} \\ \cline{3-14} 
\multicolumn{2}{c|}{} & MSE & MAE & MSE & MAE & MSE & MAE & MSE & MAE & MSE & MAE & MSE & MAE \\ \hline
 & HL  & 0.095 & 0.200 & 0.070 & 0.173 & 0.416 & 0.378 & 0.013 & 0.062 & 0.350 & 0.384 &  0.226 & 0.300 \\ 
\multirow{3}{*}{\rotatebox{90}{\textit{E2E}}} & Ridge Reg.  & 0.077 & 0.181 & 0.058 & 0.157 & 0.174 & 0.285 & 0.007 & 0.050 & 0.253 & 0.349 & 0.171 & 0.271 \\
& DLinear  & 0.081 & 0.188 & 0.062 & 0.164 & 0.170 & 0.277 & 0.013 & 0.067 & 0.261 & 0.361 & 0.171 & 0.271\\
&iTransformer  & 0.061 & \underline{0.155} & 0.044 & 0.132 & \textbf{0.160} & \textbf{0.225} & \underline{0.006} & \underline{0.046} & 0.237 & \textbf{0.325} & \underline{0.131} & \underline{0.222} \\ \hline
\multirow{5}{*}{\rotatebox{90}{\textit{MTS}}} &TS2Vec  & 0.088 & 0.206 & 0.072 & 0.194 & 0.199 & 0.314 & 0.036 & 0.146 & 0.294 & 0.395 & OOM & OOM \\ 
&CoST  & 0.075 & 0.191 & 0.052 & 0.156 & 0.192 & 0.302 & 0.014 & 0.092 & 0.228 & 0.345 & OOM & OOM \\
&PatchTST  & 0.069 & 0.172 & 0.053 & 0.150 & 0.168 & 0.272 & 0.007 & 0.050 & 0.247 & 0.346 & OOM & OOM \\
&T-Rep  & 0.058 & 0.156 & \underline{0.042} & \underline{0.131} & 0.179 & 0.279 & \textit{\textbf{0.005}} & \textit{\textbf{0.043}} & \textit{\textbf{0.217}} & \textit{\underline{0.326}} & OOM & OOM \\
&TimesURL  & 0.090 & 0.206 & 0.063 & 0.175 & 0.195 & 0.300 & OOM & OOM & OOM & OOM & OOM & OOM \\ \hline
\multirow{4}{*}{\rotatebox{90}{\textit{STS}}}& STEP*  & \underline{0.054} & 0.159 & 0.047 & \multicolumn{1}{c|}{0.156} & 0.171 & 0.279 & OOM & OOM & OOM & OOM & OOM & OOM \\
&ST-SSL*  & 0.083 & 0.200 & 0.059 & \multicolumn{1}{c|}{0.174} & 0.177 & 0.283 & 0.025 & 0.116 & 0.254 & 0.371 & OOM & OOM \\
&GPT-ST*  & 0.066 & 0.168 & 0.052 & \multicolumn{1}{c|}{0.154} & 0.170 & 0.275 & 0.012 & 0.068 & 0.239 & 0.343 & OOM & OOM \\
 &\textbf{ST-ReP}  & \textit{\textbf{0.044}} & \textit{\textbf{0.134}} & \textit{\textbf{0.033}} & \textit{\textbf{0.120}} & \textit{\underline{0.167}} & \textit{\underline{0.267}} & \textit{\textbf{0.005}} & \underline{0.046} & \underline{0.226} &0.335& \textit{\textbf{0.071}}& \textit{\textbf{0.182}}\\ \hline
\multicolumn{14}{l}{\begin{tabular}[l]{@{}l@{}}* For self-supervised-based STS forecasting methods, we only use their pre-trained encoders to generate spatial-temporal\\representations, which are then evaluated for predictive performance using the same downstream model.\end{tabular}}
\end{tabular}%

\caption{Comparison of the downstream spatial-temporal forecasting accuracy on diverse datasets. The best results are shown in bold, while the second-best results are underlined. Italicized numbers represent the best results among the self-supervised representation learning methods. OOM indicates that the method encountered an out-of-memory issue during either the pre-training or the downstream task. \textit{E2E} denotes end-to-end methods.}
\label{table_prediction}
\end{table*}

\paragraph{Datasets}
Six datasets from three different domains are used for the experiments. (1) Transportation domain: PEMS04, PEMS08 \cite{ASTGCN,STSGCN}, and CA \cite{LargeST}. (2) Climate domain: Temperature, Huimidity \cite{weatherbench}. (3) Energy domain: SDWPF \cite{SDWPF}. 
Table \ref{table_dataset_details} summarises all datasets.
  More details and setups about datasets are provided in the Appendix. Additionally, we measure the average coefficient of variation (CV) \cite{metric_CV} for each dataset, which is the ratio of the standard deviation to the mean of each variable, reflecting the variability of the data across all time points. Datasets with high CV values are typically more challenging to predict, whereas the performance gains on datasets with extremely low CV values tend to be minimal or limited.
  
\paragraph{Baselines and Implementation}
We choose twelve methods as baselines for comparison with ST-ReP: (1) Two naive methods:  \textbf{History Last (HL)} \cite{HL}, which uses the last observation as all prediction. And the \textbf{Ridge Regression} \cite{ridge_regression} model providing simple yet effective baseline.
    (2) Two end-to-end baselines: \textbf{DLinear} \cite{DLinear}, \textbf{iTransformer} \cite{iTransformer}.
    (3) Five advanced self-supervised representation learning models for MTS: \textbf{TS2Vec} \cite{TS2Vec}, \textbf{CoST} \cite{CoST}, \textbf{T-Rep} \cite{T-Rep}, \textbf{TimesURL} \cite{TimesURL} and \textbf{PatchTST} \cite{PatchTST}. 
    (4) Three self-supervised based ST forecasting model: \textbf{ST-SSL} \cite{ST-SSL}, \textbf{STEP} \cite{STEP} and \textbf{GPT-ST} \cite{GPTST}. 
     All experiments are conducted on a Linux server with one Intel(R) Xeon(R) Gold 5220 CPU @ 2.20 GHz and one 32GB NVIDIA Tesla V100-SXM2 GPU card. More details about the implementation of baselines and the proposed ST-ReP are in the Appendix.

\paragraph{Evaluation Setup}
Following the setup used in previous work \cite{TS2Vec,TimesURL}, we generate representations from $T$ historical observations using the pretrained model and predict the final $T'$ targets based on the representation $r_{:,t}$ of the most recent time step $t$.
We first train the representation learning model using the training set, and then use the learned representations and corresponding targets with various prediction length to train linear ridge regression models. The regularization weight for the ridge regression model is selected based on performance on the validation set. Finally, we evaluate the model on the test set using two common-used metrics, MSE and MAE, to measure the predictive performance. We use normalized metrics to handle the large differences in raw value scales across domains and facilitate unified presentation of results. The normalization is applied to all variables collectively, rather than individually.

Note, ridge regression does not require GPU devices or backpropagation algorithms. For training samples already converted into spatial-temporal representations, we perform sampling and use only a small fraction of the representation samples (0.93\% to 5.5\%) as training data for downstream tasks. This approach significantly reduces computational overhead, accelerates processing, and simulates resource-constrained scenarios. By contrast, for end-to-end baseline methods, a specific model needs to be retrained for each prediction horizon, using the entire training samples.

\subsection{Prediction Performance Comparison}

Table \ref{table_prediction} presents the results for a prediction horizon of 12. Results for additional prediction horizons can be found in the Appendix. Our model outperforms all self-supervised representation learning methods in most cases and even exceeds the performance of end-to-end models on certain datasets.
Specifically, on two PEMS datasets from the transportation domain, ST-ReP achieves an average improvement of 19.97\%/11.25\% in terms of MSE/MAE compared to the second best self-supervised baselines. For SDWPF dataset in energy domain, the improvement is 0.60\%/1.87\%. On two climate datasets, ST-ReP ranks second among self-supervised methods, with only T-ReP showing marginally better performance. This is likely due to the stronger influence of trends compared to spatial dependencies in these datasets, which are more effectively captured by the temporal module in T-ReP.
These results demonstrate the superior predictive performance of ST-ReP and its robust adaptability across different domain datasets with various CV values, validating its ability to learn representations that are beneficial for downstream prediction tasks. 

\begin{table}[t]
\centering
\begin{tabular}{c|cc|cc|cc}
\hline
\multirow{3}{*}{\textbf{methods}} & \multicolumn{2}{c|}{\textbf{PEMS04}} & \multicolumn{2}{c|}{\textbf{Humidity}} & \multicolumn{2}{c}{\textbf{CA}} \\ 
& \multicolumn{2}{c|}{($N$=307)} & \multicolumn{2}{c|}{($N$=2048)} & \multicolumn{2}{c}{($N$=8600)} \\ 
\cline{2-7} 
 & Fp & $T_\text{trn}$ & Fp& $T_\text{trn}$ & Fp & $T_\text{trn}$ \\ \hline
ST-SSL & 2.87 & 28.6 & 9.76 & 87.4 & \multicolumn{2}{c}{OOM} \\
GPT-ST & 4.79 & 34.2 & 23.58 & 50.7 & \multicolumn{2}{c}{OOM}  \\
STReP & \textbf{1.57} & \textbf{24.6} & \textbf{6.12} & \textbf{25.5} &\textbf{23.58}& \textbf{458.5} \\ \hline
\end{tabular}%
\caption{Comparison of computational efficiency in the pre-training stage. Fp: GPU memory footprints (GB), $\text{T}_{\text{trn}}$: the training time (s/epoch).} 
\label{table_efficiency}
\end{table}

\subsection{Efficiency Analysis}
We compare the computational efficiency of ST-ReP with other STS forecasting models, as they all include sequences of all variables within a single batch. The results are presented in Table \ref{table_efficiency}. The batch size is fixed to 32.
STEP is excluded from this comparison because its pre-training requires additional input of historical data that is hundreds of times longer, resulting in significantly higher overhead with the same batch size, making it incomparable.

The results indicate that our model has the smallest memory footprints and the shortest training time on Humidity and PEMS04 datasets. This efficiency is attributed to the linear structure of our ST-Encoder. Moreover, the representations learned by our ST-ReP are significantly lower in dimensionality. Compared to other models, which often learn representations with similar or higher dimensions (e.g., $d$d is 64 in both ST-SSL and ST-ReP, and 320 in four contrastive-based methods), the representations learned by ST-ReP achieves better predictive performance. This highlights that our ST representations have higher semantic density, reducing potential redundancy during the encoding process.

Moreover, on the CA dataset with the largest number of spatial nodes, all self-supervised learning baseline methods encounter OOM issues during both the pretraining and downstream prediction stages. This is due to their large footprint or feature dimensions. In contrast, ST-ReP is able to run on the CA dataset without reducing the model size, demonstrating its good scalability.

\begin{figure}[t]
	\centering
	\includegraphics[width=\linewidth]{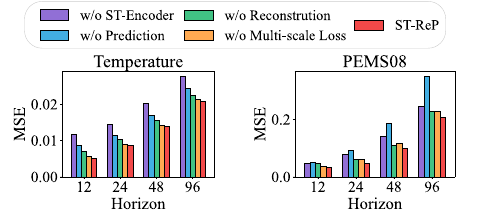}
	\caption{The results of ablation study.}
        \label{fig_ablation}
\end{figure}

\subsection{Ablation Study}
To validate the roles of key components in the model, we conduct a set of ablation experiments, setting up the following four variant models:
\textbf{(1) w/o ST-Encoder}: The ST-Encoder part of ST-ReP is removed, meaning that the ST representation generated by the model directly degenerates to $\mathbf E$ from Equation \ref{equ_e}. This model only incorporates static ST contextual information to generate future values.
\textbf{(2) w/o Prediction} and \textbf{w/o Reconstruction}: The $Decoder_{pred}$ and $Decoder_{recon}$, along with their respective loss constraints, are removed separately to assess the roles of each decoder.
\textbf{(3) w/o Multi-scale Loss}: The loss $\mathcal L _{MS}$ for multiple time scales supervision  is removed to evaluate the benefits brought by multiscale temporal analysis.

Figure \ref{fig_ablation} presents the results of ablation study. Overall, each component of ST-ReP contributed to improving its predictive performance across datasets from different domains, indicating the benefits of the ST-Encoder and the three-part loss function. 
Additionally, we find that even without the ST-Encoder, the model's predictions remained competitive with the best baseline, suggesting that the static context embedding and the designed loss function provide strong supervision. The performance degradation in the experiments where the decoder components were removed demonstrates the importance of these components in supervising the learned representations.

\section{Conclusion}
In this paper, we proposed a spatial-temporal representation learning methods ST-ReP for spatial-temporal forecasting, integrating the reconstruction for current value and prediction for unseen future value in a uniform pre-training framework. These two components, combined with multi-scale temporal analysis, constitute a ternary loss that guides the model in learning STS representations with robust predictive performance from various perspectives. Furthermore, we propose a lightweight STS encoding mechanism, build a Compression-Extraction-Decompression architecture with linear complexity. Such an architecture significantly reduces model overhead without compromising performance.
Empirical results demonstrate that the model learns representations that are both semantically dense and predictive, outperforming the best baseline methods in terms of accuracy and scalability.
This highlights ST-ReP's potential to generate compact spatial-temporal representations that can be flexibly and efficiently applied in resource-constrained downstream scenarios. We believe that this capability is an important area for STS community to explore, complementing the emphasis on accuracy improvements achieved with large-scale data and complex downstream models.
Our future work will explore fusion strategies between various spatial-temporal mining models and self-supervised learning methods to enhance representation robustness and generalization across diverse downstream tasks.

\section{Acknowledgments}
This work was partly supported by the National Key Research and Development Program of China under Grant 2022YFB4501704, the National Natural Science Foundation of China under Grant 72342026, and Fundamental Research Funds for the Central Universities under Grant 2024-6-ZD-02.
\bibliography{ST-ReP}

\begin{thebibliography}{37}
\providecommand{\natexlab}[1]{#1}

\bibitem[{Brown(1998)}]{metric_CV}
Brown, C.~E. 1998.
\newblock \emph{Coefficient of Variation}, 155--157.
\newblock Berlin, Heidelberg: Springer Berlin Heidelberg.
\newblock ISBN 978-3-642-80328-4.

\bibitem[{Challu et~al.(2023)Challu, Olivares, Oreshkin, Ramirez, Canseco, and Dubrawski}]{NHITS}
Challu, C.; Olivares, K.~G.; Oreshkin, B.~N.; Ramirez, F.~G.; Canseco, M.~M.; and Dubrawski, A. 2023.
\newblock {NHITS}: {Neural} {Hierarchical} {Interpolation} for {Time} {Series} {Forecasting}.
\newblock \emph{Proceedings of the AAAI Conference on Artificial Intelligence}, 37(6): 6989--6997.
\newblock Number: 6.

\bibitem[{Chen et~al.(2023)Chen, Chen, Shang, Wu, Zheng, Wen, and Zhang}]{MAGNN}
Chen, L.; Chen, D.; Shang, Z.; Wu, B.; Zheng, C.; Wen, B.; and Zhang, W. 2023.
\newblock Multi-{Scale} {Adaptive} {Graph} {Neural} {Network} for {Multivariate} {Time} {Series} {Forecasting}.
\newblock \emph{IEEE Transactions on Knowledge and Data Engineering}, 35(10): 10748--10761.
\newblock Conference Name: IEEE Transactions on Knowledge and Data Engineering.

\bibitem[{Choromanski et~al.(2021)Choromanski, Likhosherstov, Dohan, Song, Gane, Sarlos, Hawkins, Davis, Mohiuddin, Kaiser, Belanger, Colwell, and Weller}]{Performer}
Choromanski, K.~M.; Likhosherstov, V.; Dohan, D.; Song, X.; Gane, A.; Sarlos, T.; Hawkins, P.; Davis, J.~Q.; Mohiuddin, A.; Kaiser, L.; Belanger, D.~B.; Colwell, L.~J.; and Weller, A. 2021.
\newblock Rethinking Attention with Performers.
\newblock In \emph{International Conference on Learning Representations}.

\bibitem[{Dong et~al.(2023)Dong, Wu, Zhang, Zhang, Wang, and Long}]{SimMTM}
Dong, J.; Wu, H.; Zhang, H.; Zhang, L.; Wang, J.; and Long, M. 2023.
\newblock {SimMTM}: {A} {Simple} {Pre}-{Training} {Framework} for {Masked} {Time}-{Series} {Modeling}.
\newblock \emph{Advances in Neural Information Processing Systems}, 36: 29996--30025.

\bibitem[{Fraikin, Bennetot, and Allassonniere(2024)}]{T-Rep}
Fraikin, A.~F.; Bennetot, A.; and Allassonniere, S. 2024.
\newblock T-Rep: Representation Learning for Time Series using Time-Embeddings.
\newblock In \emph{The Twelfth International Conference on Learning Representations}.

\bibitem[{Guo et~al.(2019)Guo, Lin, Feng, Song, and Wan}]{HL}
Guo, S.; Lin, Y.; Feng, N.; Song, C.; and Wan, H. 2019.
\newblock Attention Based Spatial-Temporal Graph Convolutional Networks for Traffic Flow Forecasting.
\newblock \emph{Proceedings of the AAAI Conference on Artificial Intelligence}, 33(01): 922--929.

\bibitem[{Guo et~al.(2023)Guo, Lin, Gong, Wang, Zhou, Shen, Huang, and Wan}]{SSTBAN}
Guo, S.; Lin, Y.; Gong, L.; Wang, C.; Zhou, Z.; Shen, Z.; Huang, Y.; and Wan, H. 2023.
\newblock Self-{Supervised} {Spatial}-{Temporal} {Bottleneck} {Attentive} {Network} for {Efficient} {Long}-term {Traffic} {Forecasting}.
\newblock In \emph{2023 {IEEE} 39th {International} {Conference} on {Data} {Engineering} ({ICDE})}, 1585--1596.
\newblock ISSN: 2375-026X.

\bibitem[{Hendrycks and Gimpel(2016)}]{GELU}
Hendrycks, D.; and Gimpel, K. 2016.
\newblock Gaussian error linear units (gelus).
\newblock \emph{arXiv}, abs/1606.08415: 1--10.

\bibitem[{Huber(1992)}]{HuberLoss}
Huber, P.~J. 1992.
\newblock Robust Estimation of a Location Parameter.
\newblock In \emph{Breakthroughs in Statistics: Methodology and Distribution}, 492--518. New York, NY: Springer New York.
\newblock ISBN 978-1-4612-4380-9.

\bibitem[{Ji et~al.(2023)Ji, Wang, Huang, Wu, Xu, Wu, Zhang, and Zheng}]{ST-SSL}
Ji, J.; Wang, J.; Huang, C.; Wu, J.; Xu, B.; Wu, Z.; Zhang, J.; and Zheng, Y. 2023.
\newblock Spatio-temporal self-supervised learning for traffic flow prediction.
\newblock In \emph{Proceedings of the AAAI conference on artificial intelligence}, volume~37, 4356--4364.

\bibitem[{Li et~al.(2023)Li, Xia, Xu, and Huang}]{GPTST}
Li, Z.; Xia, L.; Xu, Y.; and Huang, C. 2023.
\newblock Generative Pre-Training of Spatio-Temporal Graph Neural Networks.
\newblock In \emph{Thirty-seventh Conference on Neural Information Processing Systems}.

\bibitem[{Liu and Chen(2024)}]{TimesURL}
Liu, J.; and Chen, S. 2024.
\newblock {TimesURL}: {Self}-{Supervised} {Contrastive} {Learning} for {Universal} {Time} {Series} {Representation} {Learning}.
\newblock \emph{Proceedings of the AAAI Conference on Artificial Intelligence}, 38(12): 13918--13926.
\newblock Number: 12.

\bibitem[{Liu et~al.(2022)Liu, Yu, Liao, Li, Lin, Liu, and Dustdar}]{Pyraformer}
Liu, S.; Yu, H.; Liao, C.; Li, J.; Lin, W.; Liu, A.~X.; and Dustdar, S. 2022.
\newblock Pyraformer: Low-Complexity Pyramidal Attention for Long-Range Time Series Modeling and Forecasting.
\newblock In \emph{International Conference on Learning Representations}.

\bibitem[{Liu et~al.(2023)Liu, Xia, Liang, Hu, Wang, BAI, Huang, Liu, Hooi, and Zimmermann}]{LargeST}
Liu, X.; Xia, Y.; Liang, Y.; Hu, J.; Wang, Y.; BAI, L.; Huang, C.; Liu, Z.; Hooi, B.; and Zimmermann, R. 2023.
\newblock Large{ST}: A Benchmark Dataset for Large-Scale Traffic Forecasting.
\newblock In \emph{Thirty-seventh Conference on Neural Information Processing Systems Datasets and Benchmarks Track}.

\bibitem[{Liu et~al.(2024)Liu, Hu, Zhang, Wu, Wang, Ma, and Long}]{iTransformer}
Liu, Y.; Hu, T.; Zhang, H.; Wu, H.; Wang, S.; Ma, L.; and Long, M. 2024.
\newblock iTransformer: Inverted Transformers Are Effective for Time Series Forecasting.
\newblock In \emph{The Twelfth International Conference on Learning Representations}.

\bibitem[{Loshchilov and Hutter(2017)}]{AdamW}
Loshchilov, I.; and Hutter, F. 2017.
\newblock Fixing Weight Decay Regularization in Adam.
\newblock \emph{CoRR}, abs/1711.05101.

\bibitem[{McDonald(2009)}]{ridge_regression}
McDonald, G.~C. 2009.
\newblock Ridge regression.
\newblock \emph{Wiley Interdisciplinary Reviews: Computational Statistics}, 1(1): 93--100.

\bibitem[{Nie et~al.(2023)Nie, Nguyen, Sinthong, and Kalagnanam}]{PatchTST}
Nie, Y.; Nguyen, N.~H.; Sinthong, P.; and Kalagnanam, J. 2023.
\newblock A Time Series is Worth 64 Words: Long-term Forecasting with Transformers.
\newblock In \emph{The Eleventh International Conference on Learning Representations}.

\bibitem[{Rasp et~al.(2020)Rasp, Dueben, Scher, Weyn, Mouatadid, and Thuerey}]{weatherbench}
Rasp, S.; Dueben, P.~D.; Scher, S.; Weyn, J.~A.; Mouatadid, S.; and Thuerey, N. 2020.
\newblock {WeatherBench}: {A} benchmark dataset for data-driven weather forecasting.
\newblock \emph{Journal of Advances in Modeling Earth Systems}, 12(11): e2020MS002203.
\newblock ArXiv:2002.00469 [physics, stat].

\bibitem[{Shabani et~al.(2023)Shabani, Abdi, Meng, and Sylvain}]{Scaleformer}
Shabani, M.~A.; Abdi, A.~H.; Meng, L.; and Sylvain, T. 2023.
\newblock Scaleformer: Iterative Multi-scale Refining Transformers for Time Series Forecasting.
\newblock In \emph{The Eleventh International Conference on Learning Representations}.

\bibitem[{Shao et~al.(2022)Shao, Zhang, Wang, and Xu}]{STEP}
Shao, Z.; Zhang, Z.; Wang, F.; and Xu, Y. 2022.
\newblock Pre-training {Enhanced} {Spatial}-temporal {Graph} {Neural} {Network} for {Multivariate} {Time} {Series} {Forecasting}.
\newblock In \emph{Proceedings of the 28th {ACM} {SIGKDD} {Conference} on {Knowledge} {Discovery} and {Data} {Mining}}, {KDD} '22, 1567--1577. New York, NY, USA: Association for Computing Machinery.
\newblock ISBN 978-1-4503-9385-0.

\bibitem[{Song et~al.(2020{\natexlab{a}})Song, Lin, Guo, and Wan}]{ASTGCN}
Song, C.; Lin, Y.; Guo, S.; and Wan, H. 2020{\natexlab{a}}.
\newblock Spatial-Temporal Synchronous Graph Convolutional Networks: A New Framework for Spatial-Temporal Network Data Forecasting.
\newblock In \emph{Proceedings of the AAAI Conference on Artificial Intelligence}, volume~34, 914--921.

\bibitem[{Song et~al.(2020{\natexlab{b}})Song, Lin, Guo, and Wan}]{STSGCN}
Song, C.; Lin, Y.; Guo, S.; and Wan, H. 2020{\natexlab{b}}.
\newblock Spatial-Temporal Synchronous Graph Convolutional Networks: A New Framework for Spatial-Temporal Network Data Forecasting.
\newblock In \emph{Proceedings of the AAAI Conference on Artificial Intelligence}, volume~34, 914--921.

\bibitem[{Wang et~al.(2024)Wang, Wu, Shi, Hu, Luo, Ma, Zhang, and ZHOU}]{TimeMixer}
Wang, S.; Wu, H.; Shi, X.; Hu, T.; Luo, H.; Ma, L.; Zhang, J.~Y.; and ZHOU, J. 2024.
\newblock TimeMixer: Decomposable Multiscale Mixing for Time Series Forecasting.
\newblock In \emph{The Twelfth International Conference on Learning Representations}.

\bibitem[{Woo et~al.(2022)Woo, Liu, Sahoo, Kumar, and Hoi}]{CoST}
Woo, G.; Liu, C.; Sahoo, D.; Kumar, A.; and Hoi, S. 2022.
\newblock Co{ST}: Contrastive Learning of Disentangled Seasonal-Trend Representations for Time Series Forecasting.
\newblock In \emph{International Conference on Learning Representations}.

\bibitem[{Wu et~al.(2023{\natexlab{a}})Wu, Hu, Liu, Zhou, Wang, and Long}]{TimesNet}
Wu, H.; Hu, T.; Liu, Y.; Zhou, H.; Wang, J.; and Long, M. 2023{\natexlab{a}}.
\newblock TimesNet: Temporal 2D-Variation Modeling for General Time Series Analysis.
\newblock In \emph{The Eleventh International Conference on Learning Representations}.

\bibitem[{Wu et~al.(2023{\natexlab{b}})Wu, Yang, Zhao, He, Wipf, and Yan}]{Difformer}
Wu, Q.; Yang, C.; Zhao, W.; He, Y.; Wipf, D.; and Yan, J. 2023{\natexlab{b}}.
\newblock {DIFF}ormer: Scalable (Graph) Transformers Induced by Energy Constrained Diffusion.
\newblock In \emph{The Eleventh International Conference on Learning Representations}.

\bibitem[{Yue et~al.(2022)Yue, Wang, Duan, Yang, Huang, Tong, and Xu}]{TS2Vec}
Yue, Z.; Wang, Y.; Duan, J.; Yang, T.; Huang, C.; Tong, Y.; and Xu, B. 2022.
\newblock {TS2Vec}: {Towards} {Universal} {Representation} of {Time} {Series}.
\newblock \emph{Proceedings of the AAAI Conference on Artificial Intelligence}, 36(8): 8980--8987.
\newblock Number: 8.

\bibitem[{Zeng et~al.(2023)Zeng, Chen, Zhang, and Xu}]{DLinear}
Zeng, A.; Chen, M.; Zhang, L.; and Xu, Q. 2023.
\newblock Are transformers effective for time series forecasting?
\newblock In \emph{Proceedings of the Thirty-Seventh AAAI Conference on Artificial Intelligence and Thirty-Fifth Conference on Innovative Applications of Artificial Intelligence and Thirteenth Symposium on Educational Advances in Artificial Intelligence}, AAAI'23/IAAI'23/EAAI'23. AAAI Press.
\newblock ISBN 978-1-57735-880-0.

\bibitem[{Zerveas et~al.(2021)Zerveas, Jayaraman, Patel, Bhamidipaty, and Eickhoff}]{TST}
Zerveas, G.; Jayaraman, S.; Patel, D.; Bhamidipaty, A.; and Eickhoff, C. 2021.
\newblock A {Transformer}-based {Framework} for {Multivariate} {Time} {Series} {Representation} {Learning}.
\newblock In \emph{Proceedings of the 27th {ACM} {SIGKDD} {Conference} on {Knowledge} {Discovery} \& {Data} {Mining}}, {KDD} '21, 2114--2124. New York, NY, USA: Association for Computing Machinery.
\newblock ISBN 978-1-4503-8332-5.

\bibitem[{Zhang et~al.(2024{\natexlab{a}})Zhang, Wen, Zhang, Zheng, Li, and Bian}]{metric_Strength}
Zhang, J.; Wen, X.; Zhang, Z.; Zheng, S.; Li, J.; and Bian, J. 2024{\natexlab{a}}.
\newblock Prob{TS}: Benchmarking Point and Distributional Forecasting across Diverse Prediction Horizons.
\newblock In \emph{The Thirty-eight Conference on Neural Information Processing Systems Datasets and Benchmarks Track}.

\bibitem[{Zhang et~al.(2024{\natexlab{b}})Zhang, Wen, Zhang, Cai, Jin, Liu, Zhang, Liang, Pang, Song, and Pan}]{survey_ssl4ts}
Zhang, K.; Wen, Q.; Zhang, C.; Cai, R.; Jin, M.; Liu, Y.; Zhang, J.~Y.; Liang, Y.; Pang, G.; Song, D.; and Pan, S. 2024{\natexlab{b}}.
\newblock Self-{Supervised} {Learning} for {Time} {Series} {Analysis}: {Taxonomy}, {Progress}, and {Prospects}.
\newblock \emph{IEEE Transactions on Pattern Analysis and Machine Intelligence}, 1--20.
\newblock Conference Name: IEEE Transactions on Pattern Analysis and Machine Intelligence.

\bibitem[{Zhang and Yan(2023)}]{Crossformer}
Zhang, Y.; and Yan, J. 2023.
\newblock Crossformer: Transformer Utilizing Cross-Dimension Dependency for Multivariate Time Series Forecasting.
\newblock In \emph{The Eleventh International Conference on Learning Representations}.

\bibitem[{Zheng and Zhang(2023)}]{TAGnn}
Zheng, Q.; and Zhang, Y. 2023.
\newblock TAGnn: Time Adjoint Graph Neural Network for Traffic Forecasting.
\newblock In \emph{Database Systems for Advanced Applications: 28th International Conference, DASFAA 2023, Tianjin, China, April 17–20, 2023, Proceedings, Part I}, 369--379. Springer-Verlag.
\newblock ISBN 978-3-031-30636-5.

\bibitem[{Zhou et~al.(2021)Zhou, Zhang, Peng, Zhang, Li, Xiong, and Zhang}]{Informer}
Zhou, H.; Zhang, S.; Peng, J.; Zhang, S.; Li, J.; Xiong, H.; and Zhang, W. 2021.
\newblock Informer: {Beyond} {Efficient} {Transformer} for {Long} {Sequence} {Time}-{Series} {Forecasting}.
\newblock \emph{Proceedings of the AAAI Conference on Artificial Intelligence}, 35(12): 11106--11115.
\newblock Number: 12.

\bibitem[{Zhou et~al.(2022)Zhou, Lu, Xiao, Su, Lyu, Ma, and Dou}]{SDWPF}
Zhou, J.; Lu, X.; Xiao, Y.; Su, J.; Lyu, J.; Ma, Y.; and Dou, D. 2022.
\newblock SDWPF: A Dataset for Spatial Dynamic Wind Power Forecasting Challenge at KDD Cup 2022.
\newblock arXiv:2208.04360.

\end{thebibliography}

\clearpage 

\appendix
 \section{Appendix A: Experiments Setup Details}

 \begin{table*}[t]
\centering
\begin{tabular}{c|c|c|c|c|c|c|c}
\hline
\textbf{Dataset} & \textbf{domain}& \textbf{\#Nodes} & \textbf{\#Samples} & \textbf{\begin{tabular}[c]{@{}c@{}}Time\\ Interval\end{tabular}} & \textbf{CV} & \textbf{\begin{tabular}[c]{@{}c@{}}Trend\\ Strength\end{tabular}} & \textbf{\begin{tabular}[c]{@{}c@{}}Seasonality\\ Strength\end{tabular}}  \\
\hline
PEMS04          & transportation & 307  & 16992 & 5min   & 58.82 & 0.4315 & 0.9827\\
PEMS08         & transportation  & 170  & 17856 & 5min   & 46.75 & 0.4630 & 0.9813\\
CA             & transportation  & 8600 & 35040 & 15min  & 60.10 & 0.3392 & 0.9209\\
SDWPF          & energy & 134  & 35280 & 10min  & 121.97 & 0.5723 & 0.5430\\
Humidity      &climate  & 2048 & 8784  & 1hour  & 17.19 & 0.7508 & 0.4499\\
Temperature   &climate & 2048 & 8784  & 1hour  & 2.19  & 0.8968 & 0.5561\\
\hline
\end{tabular}

\caption{Details of datasets. CV denotes the coefficient of variation.}
\label{table_dataset_details_full}
\end{table*}

\paragraph{Details for Benchmark Datasets}
Six datasets from three different domains are used for the experiments: 
\begin{enumerate}
    \item Transportation domain: PEMS04, PEMS08 \cite{ASTGCN,STSGCN}, and CA \cite{LargeST}. The variables of these three datasets are detectors on the road network from California. They all record the highway traffic status. We use the traffic flow as the investigated feature. For CA, which owns a large size of variables, we use data from the year 2019.
    \item Climate domain: Temperature, Huimidity. These two datasets are provided in WeatherBench \cite{weatherbench}. The variables for these two datasets represent the Earth's surface, divided into 2048 grid areas (i.e., $32\times 64$). Each grid records weather information such as  or Humidity. One year of data from 2016 is used for our experiments.
    \item Energy domain: SDWPF \cite{SDWPF}. This dataset records the power generation of wind turbines over 245 days in a wind farm. The variables are wind turbines.
\end{enumerate}
  Table \ref{table_dataset_details_full} provides details about datasets. Additionally, we measure the average coefficient of variation (CV) \cite{metric_CV} for each dataset, which is the ratio of the standard deviation to the mean of each variable, reflecting the variability of the data across all time points. Datasets with high CV values are typically more challenging to predict, whereas the performance gains on datasets with extremely low CV values tend to be minimal. Among the six datasets, Temperature has the lowest CV value (2.19), but it means that the improvement space of its prediction accuracy may be limited, while SDWPF has the highest CV value.
  The trend strength and seasonality strength in the last two columns are two metics introduced in a recent study \cite{metric_Strength}.
They quantify the strengths of trend and seasonality for a fixed-length time-series segment. The values range from 0 to 1, with higher values indicating stronger strength.

  We divide all datasets into training, validation, and test sets in a 6:2:2 ratio for both self-supervised learning and downstream forecasting tasks. All data were standardized using Z-score normalization. The normalization is applied to all variables collectively, rather than individually.
  
\paragraph{Evaluation Setup}
Following the setup used in previous work \cite{TS2Vec,TimesURL}, we generate representations from $T$ historical observations using the pretrained model and predict the final $T'$ targets based on the representation $r_{:,t}$ of the most recent time step $t$.
We first train the representation learning model using the training set, and then use the learned representations and corresponding targets with various prediction length to train linear ridge regression models. The regularization weight for the ridge regression model is selected based on performance on the validation set. Finally, we evaluated the model on the test set using two common-used metrics, MSE and MAE, to measure the predictive performance. We use normalized metrics to handle the large differences in raw value scales across domains and facilitate unified presentation of results. 

Note, ridge regression does not require GPU devices or backpropagation algorithms. For training samples already converted into spatial-temporal representations, we perform sampling and use only a small fraction of the representation samples (0.93\% to 5.5\%) as training data for downstream tasks. This approach significantly reduces computational overhead, accelerates processing, and simulates resource-constrained scenarios. By contrast, for end-to-end baseline methods, a specific model needs to be retrained for each prediction horizon, using the entire training samples.

\paragraph{Baselines and Implementation Details}
We choose twelve methods as baselines for comparison with ST-ReP:
\begin{itemize}
    \item Two naive methods:  One is 
    \textbf{History Last} (HL) \cite{HL}, which uses the last observation as all prediction. The other is \textbf{Ridge regression} model. Here we directly use the historical $T$ observations to predict the future $T'$ values. Note that the ridge regression here is consistent with the method in the downstream forecasting task. The difference lies in that here the entire original data is directly used to predict future data, while the input for the downstream tasks is the learned spatial-temporal representations. Notably, ridge regression does not require GPU devices or the backpropagation algorithm. Retraining is needed for each forecasting horizon, but the training cost is extremely small compared to deep learning methods. 
    \item Two end-to-end methods: \textbf{DLinear} \cite{DLinear} and \textbf{iTransformer} \cite{iTransformer}, two lightweight and effective models for MTS forecasting. The latter focus on extracting correlations among variables by Transformer, which is in line with the idea of spatial modeling for STS forecasting. As end-to-end methods, these two models also require retraining with the entire training set for each forecasting horizon, resulting in a relatively large overall cost. 
    \item Five advanced self-supervised representation learning models for MTS: \textbf{TS2Vec} \cite{TS2Vec}, \textbf{CoST}\cite{CoST}, \textbf{T-Rep} \cite{T-Rep}, \textbf{TimesURL} \cite{TimesURL} and \textbf{PatchTST} \cite{PatchTST}. 
    The first two methods are contrastive models that treat the time series of each variable as independent samples for training. T-Rep and TimesURL are two recent advanced methods based on the architecture of TS2Vec. 
    For PatchTST, we use its self-supervised learning version. This method is based on reconstruction and, during pre-training, also focuses on reconstructing the time series of individual variables.
    
    \item Three self-supervised based STS forecasting model: \textbf{STEP} \cite{STEP}, \textbf{ST-SSL} \cite{ST-SSL}, and \textbf{GPT-ST} \cite{GPTST}. In ST-SSL, the spatial correlations  of all variables are explicitly modeled in each batch. Since its prediction is still in the form of end-to-end forecasting, we treat it as a pre-trained model and use its learned representations for downstream task comparison. STEP is a reconstruction-based method based on a Transformer architecture that utilizes patching and long inputs (e.g., up to two weeks). It feeds the learned representations into downstream spatial-temporal neural networks to assist in spatial-temporal forecasting. We use its pre-trained encoder (referred to as TSFormer in the original paper) and evaluate the learned representations using the standardized downstream task in this experiment (ridge regression). Similarly, for GPT-ST, we pre-train it and then evaluate its learned spatial-temporal representations.
\end{itemize}
For all self-supervised methods, after completing the pre-training, we use their learned representations to perform the same downstream tasks as described in the Evaluation Setup.

All experiments are conducted on a Linux server with one Intel(R) Xeon(R) Gold 5220 CPU @ 2.20 GHz and one 32GB NVIDIA Tesla V100-SXM2 GPU card. For ST-ReP, the input feature length $C$ is $1$ for all datasets. The input length $T$ is $12$, the length of targeted future value $F$ is set to $12$, and the predicted length $T'$ in the downstream tasks varies in $\{4,8,12,16\}$ for CA and $\{12,24,48,96\}$ for others. The size of time compression space $p$ is set to $3$, and the size of proxy tensor $m$ in spatial extraction is $8$. The representation dimension $d$ is $64$, and the number of encoder layers $L$ are $3$. The kernel sizes of the average pooling filters $\Omega$ are $\{2,4,8,16\}$. AdamW\cite{AdamW} is used as the optimizer during the pre-training, with a learning rate of 0.001 for up to 100 epochs, and early stopping strategy is employed. The batch size is $32$. The group of weights in the loss function $\{\alpha,\beta\}$are searched over $\{0.1,0.2,0.3,0.4,0.5\}$.
The hyper-parameters are set based on the best performance on the validation set.
For downstream forecasting, the regularization weight in the ridge regression are searched over$\{0.1,0.2,0.5,1,2,5,10,20,50,100,200,500,1000\}$. We report the average metrics over ten repetitions as results.

For the four self-supervised MTS methods (TS2Vec, CoST, T-Rep, TimesURL), the representation dimension $d$ is $320$ and the batch size is $8$. For the remaining models, $d$ is $64$ and the batch size is $32$, (except that $d$ for PatchTST is $128$ and the batch size of iTransformer on the CA dataset is $4$).

\section{Appendix B: Experiment results}

\begin{table*}[]
\centering
\begin{tabular}{c|c|cc|cc|cc|cc|cc|c}
\hline
\multirow{2}{*}{\textbf{Methods}} & \multirow{2}{*}{\textbf{horizon}} & \multicolumn{2}{c|}{\textbf{PEMS04}} & \multicolumn{2}{c|}{\textbf{PEMS08}} & \multicolumn{2}{c|}{\textbf{SDWPF}} & \multicolumn{2}{c|}{\textbf{temperature}} & \multicolumn{2}{c|}{\textbf{humidity}} & \multirow{2}{*}{\begin{tabular}[c]{@{}c@{}}\textbf{Avg}\\ \textbf{Rank}\end{tabular}} \\
\cline{3-12} 
 &  & MSE & MAE & MSE & MAE & MSE & MAE & MSE & MAE & MSE & MAE & \multicolumn{1}{l}{} \\ \hline
\multirow{4}{*}{\begin{tabular}[c]{@{}c@{}}HL\\ (Naive)\end{tabular}} & 12 & 0.095 & 0.200 & 0.070 & 0.173 & 0.416 & 0.378 & 0.013 & 0.062 & 0.350 & 0.384 &  \\
 & 24 & 0.181 & 0.279 & 0.144 & 0.248 & 0.650 & 0.498 & 0.016 & 0.074 & 0.477 & 0.470 &  \\
 & 48 & 0.415 & 0.433 & 0.395 & 0.351 & 0.966 & 0.642 & 0.022 & 0.090 & 0.593 & 0.540 &  \\
 & 96 & 0.919 & 0.683 & 0.805 & 0.631 & 1.270 & 0.771 & 0.030 & 0.107 & 0.679 & 0.589 &  \\ \hline
\multirow{4}{*}{\begin{tabular}[c]{@{}c@{}}Ridge Reg.\\ (Naive)\end{tabular}} & 12 & 0.077 & 0.181 & 0.058 & 0.157 & 0.174 & 0.285 & 0.007 & 0.050 & 0.253 & 0.349 &  \\
 & 24 & 0.145 & 0.253 & 0.115 & 0.224 & 0.290 & 0.387 & 0.010 & 0.061 & 0.335 & 0.418 &  \\
 & 48 & 0.310 & 0.386 & 0.261 & 0.354 & 0.468 & 0.514 & 0.015 & 0.076 & 0.405 & 0.471 &  \\
 & 96 & 0.582 & 0.562 & 0.533 & 0.542 & 0.662 & 0.630 & 0.023 & 0.093 & 0.454 & 0.507 &  \\ \hline
\multirow{4}{*}{\begin{tabular}[c]{@{}c@{}}DLinear\\ (End-to-end)\end{tabular}} & 12 & 0.081 & 0.188 & 0.062 & 0.164 & 0.170 & 0.277 & 0.013 & 0.067 & 0.261 & 0.361 &  \\
 & 24 & 0.149 & 0.259 & 0.120 & 0.233 & 0.278 & 0.373 & 0.014 & 0.072 & 0.338 & 0.422 &  \\
 & 48 & 0.314 & 0.390 & 0.268 & 0.362 & 0.457 & 0.500 & 0.018 & 0.085 & 0.406 & 0.473 &  \\
 & 96 & 0.586 & 0.564 & 0.537 & 0.546 & \textbf{0.653} & \underline{0.616} & 0.025 & 0.100 & 0.454 & 0.508 &  \\ \hline
\multirow{4}{*}{\begin{tabular}[c]{@{}c@{}}iTransformer\\ (End-to-end)\end{tabular}} & 12 & 0.061 & \underline{0.155} & 0.044 & 0.132 & \textbf{0.160} & \textbf{0.225} & \underline{0.006} & \underline{0.046} & 0.237 & \textbf{0.325} &  \\
 & 24 & 0.105 & 0.203 & 0.081 & 0.178 & \textbf{0.267} & \textbf{0.301} & \underline{0.009} & \underline{0.057} & 0.335 & 0.403 &  \\
 & 48 & 0.247 & 0.313 & 0.203 & 0.288 & \textbf{0.448} & \textbf{0.403} & 0.015 & \underline{0.074} & 0.420 & 0.465 &  \\
 & 96 & 0.526 & 0.485 & 0.415 & 0.422 & 0.730 & \textbf{0.532} & 0.022 & \underline{0.091} & 0.483 & 0.508 &  \\ \hline
\multirow{4}{*}{TS2Vec} & 12 & 0.088 & 0.206 & 0.072 & 0.194 & 0.199 & 0.314 & 0.036 & 0.146 & 0.294 & 0.395 &  \\
 & 24 & 0.144 & 0.261 & 0.114 & 0.243 & 0.351 & 0.457 & 0.033 & 0.139 & 0.371 & 0.454 &  \\
 & 48 & 0.267 & 0.363 & 0.203 & 0.328 & 0.579 & 0.613 & 0.041 & 0.154 & 0.445 & 0.507 &  \\
 & 96 & 0.384 & 0.449 & 0.366 & 0.453 & 0.821 & 0.734 & 0.055 & 0.176 & 0.493 & 0.539 &  \\ \hline
\multirow{4}{*}{CoST} & 12 & 0.075 & 0.191 & 0.052 & 0.156 & 0.192 & 0.302 & 0.014 & 0.092 & 0.228 & 0.345 &  \\
 & 24 & 0.117 & 0.236 & 0.084 & 0.202 & 0.305 & 0.387 & 0.015 & 0.092 & 0.302 & 0.410 &  \\
 & 48 & 0.184 & 0.302 & 0.141 & 0.268 & 0.494 & 0.527 & 0.020 & 0.102 & \underline{0.347} & 0.446 &  \\
 & 96 & 0.243 & 0.356 & 0.191 & 0.318 & 0.708 & 0.659 & 0.025 & 0.111 & 0.391 & 0.481 &  \\ \hline
\multirow{4}{*}{T-Rep} & 12 & 0.058 & 0.156 & \underline{0.042} & \underline{0.131} & 0.179 & 0.279 & \textit{\textbf{0.005}} & \textit{\textbf{0.043}} & \textit{\textbf{0.217}} & \underline{\textit{0.326}} & \underline{2.4} \\
 & 24 & 0.089 & 0.196 & 0.063 & \underline{0.164} & 0.300 & 0.381 & \textit{\textbf{0.008}} & \textit{\textbf{0.055}} & \textit{\textbf{0.279}} & \textit{\textbf{0.384}} & 2.7 \\
 & 48 & 0.144 & 0.255 & 0.099 & 0.212 & 0.488 & 0.511 & \textit{\textbf{0.013}} & \textit{\textbf{0.070}} & \textit{\textbf{0.326}} & \textit{\textbf{0.424}} & 3 \\ 
 & 96 & \underline{0.180} & \textit{\textbf{0.289}} & \underline{0.130} & \textit{\textbf{0.245}} & 0.702 & 0.633 & \textit{\textbf{0.019}} & \textit{\textbf{0.086}} & \textit{\textbf{0.358}} & \textit{\textbf{0.450}} & \textit{\textbf{2.3}} \\ \hline
\multirow{4}{*}{TimesURL} & 12 & 0.090 & 0.206 & 0.063 & 0.175 & 0.195 & 0.300 & OOM & OOM & OOM & OOM &  \\
 & 24 & 0.132 & 0.248 & 0.120 & 0.249 & 0.306 & 0.384 & OOM & OOM & OOM & OOM &  \\
 & 48 & 0.237 & 0.342 & 0.207 & 0.336 & 0.483 & 0.497 & OOM & OOM & OOM & OOM &  \\
 & 96 & 0.428 & 0.489 & 0.968 & 0.729 & 0.698 & \textit{0.621} & OOM & OOM & OOM & OOM &  \\ \hline
\multirow{4}{*}{STEP} & 12 & \underline{0.054} & 0.159 & 0.047 & 0.156 & 0.171 & 0.279 & OOM & OOM & OOM & OOM & 3.7 \\
 & 24 & \underline{0.068} & \underline{0.181} & \underline{0.058} & 0.172 & 0.283 & 0.377 & OOM & OOM & OOM & OOM & \underline{2.5} \\
 & 48 & \textit{\textbf{0.109}} & \underline{0.233} & \textit{\textbf{0.081}} & \textit{\textbf{0.203}} & 0.465 & 0.509 & OOM & OOM & OOM & OOM & \underline{2.7} \\
 & 96 & \textit{\textbf{0.178}} & \underline{0.306} & \textit{\textbf{0.123}} & \underline{0.254} & 0.680 & 0.637 & OOM & OOM & OOM & OOM & 3.0 \\ \hline
\multirow{4}{*}{PatchTST} & 12 & 0.069 & 0.172 & 0.053 & 0.150 & 0.168 & 0.272 & 0.007 & 0.050 & 0.247 & 0.346 &  \\
 & 24 & 0.123 & 0.234 & 0.099 & 0.209 & 0.280 & 0.374 & 0.010 & 0.060 & 0.324 & 0.412 &  \\
 & 48 & 0.264 & 0.354 & 0.225 & 0.328 & \underline{0.456} & 0.504 & 0.015 & 0.076 & 0.388 & 0.463 &  \\
 & 96 & 0.519 & 0.523 & 0.471 & 0.505 & 0.664 & 0.630 & 0.022 & 0.093 & 0.433 & 0.496 &  \\ \hline
\multirow{4}{*}{ST-SSL} & 12 & 0.083 & 0.200 & 0.059 & 0.174 & 0.177 & 0.283 & 0.025 & 0.116 & 0.254 & 0.371 &  \\
 & 24 & 0.142 & 0.263 & 0.094 & 0.220 & 0.284 & 0.378 & 0.026 & 0.117 & 0.315 & 0.417 &  \\
 & 48 & 0.274 & 0.374 & 0.185 & 0.312 & \underline{0.456} & 0.504 & 0.031 & 0.127 & 0.371 & 0.458 &  \\
 & 96 & 0.427 & 0.485 & 0.328 & 0.425 & 0.660 & 0.628 & 0.038 & 0.140 & 0.411 & 0.486 &  \\ \hline
\multirow{4}{*}{GPT-ST} & 12 & 0.066 & 0.168 & 0.052 & 0.154 & 0.170 & 0.275 & 0.012 & 0.068 & 0.239 & 0.343 &  \\
 & 24 & 0.105 & 0.217 & 0.095 & 0.213 & 0.286 & 0.379 & 0.014 & 0.077 & 0.311 & 0.406 &  \\
 & 48 & 0.187 & 0.300 & 0.199 & 0.317 & 0.463 & 0.508 & 0.020 & 0.092 & 0.367 & 0.451 &  \\
 & 96 & 0.294 & 0.389 & 0.348 & 0.438 & 0.669 & 0.632 & 0.029 & 0.110 & 0.403 & 0.480 &  \\ \hline
\multirow{4}{*}{ST-ReP (ours)} & 12 & \textit{\textbf{0.044}} & \textit{\textbf{0.134}} & \textit{\textbf{0.033}} & \textit{\textbf{0.120}} & \underline{\textit{0.167}} & \underline{\textit{0.267}} & \textit{\textbf{0.005}} & \underline{0.046} & \underline{0.226} & 0.335 & \textit{\textbf{1.4}} \\
 & 24 & \textit{\textbf{0.058}} & \textit{\textbf{0.155}} & \textit{\textbf{0.048}} & \textit{\textbf{0.145}} & \underline{\textit{0.277}} & \underline{\textit{0.367}} & \underline{0.009} & 0.059 & \underline{0.296} & \underline{0.396} & \textit{\textbf{1.4}} \\
 & 48 & \underline{0.115} & \textit{\textbf{0.222}} & \underline{0.098} & \underline{0.211} & \textit{\textbf{0.448}} & \underline{\textit{0.495}} & \underline{0.014} & 0.075 & 0.352 & \underline{0.442} & \textit{\textbf{1.8}} \\
 & 96 & 0.230 & 0.329 & 0.208 & 0.322 & \underline{\textit{0.657}} & 0.624 & \underline{0.021} & \underline{0.091} & \underline{0.389} & \underline{0.471} & \underline{2.5}\\
\hline
\end{tabular}%

\caption{Full results over various horizons on five datasets. The best results are shown in bold, while the second-best results are underlined. Italicized numbers represent the best results among the self-supervised representation learning methods. OOM indicates that the method encountered an out-of-memory issue during either the pre-training or the downstream task. The last column shows the ranking of the three state-of-the-art methods among nine self-supervised methods, based on the average ranking across all datasets.}
\label{table_fullresults}
\end{table*}

\paragraph{Prediction Performance Comparison}

Table \ref{table_prediction} presents the results for a prediction horizon of 12. Results for additional prediction horizons can be found in the Appendix. Our model outperforms all self-supervised representation learning methods in most cases and even exceeds the performance of end-to-end models on certain datasets.
Specifically, on two PEMS datasets from the transportation domain, ST-ReP achieves an average improvement of 19.97\%/11.25\% in terms of MSE/MAE compared to the second best self-supervised baselines. For SDWPF dataset in energy domain, the improvement is 0.60\%/1.87\%. On two climate datasets, ST-ReP ranks second among self-supervised methods, with only T-ReP showing marginally better performance. This is likely due to the stronger influence of trends compared to spatial dependencies in these datasets, which are more effectively captured by the temporal module in T-ReP.

For the longer prediction horizons of 48 and 96, ST-ReP performs worse than STEP. A possible reason is that STEP uses a longer input window (2 weeks) during pre-training, making the model more inclined to retain long-term data insights. However, compared to other models that use the same input length (12), ST-ReP still maintains superior predictive performance.
Among nine self-supervised based methods, ST-ReP achieves the highest average ranking, these results demonstrate the superior predictive performance of ST-ReP and its robust adaptability across different domain datasets, validating its ability to learn representations that are beneficial for downstream prediction tasks. In addition, we note that further research is needed regarding the impact of input and output length settings in pre-training on downstream forecasting results, which is also part of our future work.

HL provides a basic benchmark for predictions, reflecting the intrinsic statistical characteristics of the dataset. Similarly, ridge regression, using a naive model for end-to-end predictions, benefits from abundant data sources to deliver solid baseline results. However, it is significantly affected by data noise and lacks effective generalization capabilities. 
TS2Vec and CoST focus solely on temporal modeling without considering spatial relationships between variables. Their approach of selecting other variables as negative examples introduces misleading information for contrastive learning, resulting in poor performance in downstream tasks on ST datasets. T-Rep introduces new pretext losses to address the binary limitations in contrastive learning methods, aiding the model in expressing how features are similar. It shows promising performance in long-term forecasting and on datasets with stronger temporal characteristics (such as climate datasets), thus achieving the second-best average ranking. 
TimesURL, on the other hand, introduces a new frequency-temporal-based augmentation method to help address the construction of positive and negative pairs. However, experimental results indicate that neither method achieves significant progress when applied to STS data, highlighting the challenges of modeling STS data.

Another contrastive method, ST-SSL, incorporates spatial relationships and topological augmentations to construct discriminative spatiotemporal representations. Because of its explicit consideration of spatial relationships, it achieves improvements on most datasets. In contrast, ST-ReP does not use a contrastive paradigm, thereby avoiding the burden of augmentations and the complexities of designing positive and negative pairs. Instead, it employs a reconstruction-based method, achieving the best results. 

Three reconstruction-based methods (PatchTST, STEP, GPT-ST), achieve competitive prediction accuracy on most datasets, indicating the benefits of generative methods for understanding data. In comparison, ST-ReP incorporates a prediction component and multi-scale temporal analysis during training, enforcing the model to learn coarse-grained temporal information beneficial for long-term future predictions. The superior performance of ST-ReP demonstrates the predictive capability of the learned representations.

Moreover, on the CA dataset with the largest number of spatial nodes, all self-supervised learning baseline methods encounter out-of-memory (OOM) issues during both the pretraining and downstream prediction stages. This is due to their large footprint or feature dimensions. In contrast, ST-ReP is able to run on the CA dataset without reducing the model size, demonstrating its good scalability.

\begin{figure}[t]
	\centering
	\includegraphics[width=\linewidth]{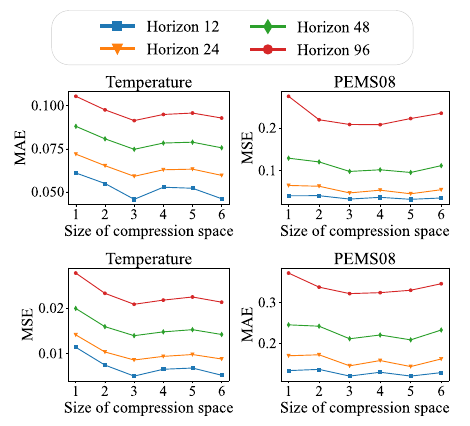}
	\caption{The results of varying size of temporal compression space $p$.}
        \label{fig_variant_tf}
\end{figure}

\begin{figure}[t]
	\centering
	\includegraphics[width=\linewidth]{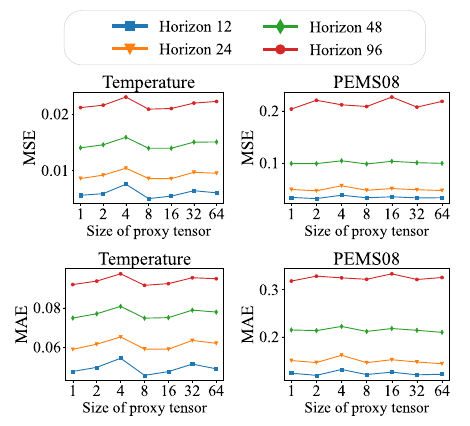}
	\caption{The results of varying size of spatial proxy tensor $m$.}
        \label{fig_variant_m}
\end{figure}
\paragraph{Hyperparameter Sensitivity Analysis}
Thanks to the temporal compression mechanism and linear spatial attention extraction, the model significantly reduces memory usage while maintaining prediction accuracy.
To explore the impact of key parameters within these two components, namely the size of the compression space $p$ and the size of the proxy tensor $m$, we  assess the model's sensitivity to these  two parameters.

Figure \ref{fig_variant_tf} and Figure \ref{fig_variant_m} illustrate the model's performance across different prediction horizons as $p$ and $m$ increase. The results show that as the $p$ value increases, the prediction error initially decreases and then increases, especially over longer time dependencies. In contrast, the model does not exhibit a strong dependence on the spatial proxy size $m$, showing no clear trend, although increasing $m$ does raise computational costs.
Setting these two parameters as constants helps reduce the model's computational cost without sacrificing much performance. This allows the model to adapt to more complex and large-scale datasets, capturing ST features within a low-dimensional space, thereby reducing redundancy and enhancing scalability.

\end{document}